# Face Detection Using Radial Basis Function Neural Networks With Fixed Spread Value

Khairul Azha A. Aziz
*Faculty of Electronics and Computer Engineering,
Universiti Teknikal Malaysia Melaka,
Ayer Keroh, Melaka, Malaysia.*
khairulazha@utem.edu.my

Shahrum Shah Abdullah
*Faculty of Electrical Engineering,
Universiti Teknologi Malaysia,
Skudai, Malaysia.*
shahrum@fke.utm.my

*Abstract*—**This paper presented a face detection system using Radial Basis Function Neural Networks With Fixed Spread Value. Face detection is the first step in face recognition system. The purpose is to localize and extract the face region from the background that will be fed into the face recognition system for identification. General preprocessing approach was used for normalizing the image and Radial Basis Function (RBF) Neural Network was used to distinguish between face and non-face. RBF Neural Networks offer several advantages compared to other neural network architecture such as they can be trained using fast two stages training algorithm and the network possesses the property of best approximation. The output of the network can be optimized by setting suitable value of center and spread of the RBF. In this paper, fixed spread value will be used. The Radial Basis Function Neural Network (RBFNN) used to distinguish faces and non-faces and the evaluation of the system will be the performance of detection, False Acceptance Rate (FAR), False Rejection Rate (FRR) and the discriminative properties.**

I. INTRODUCTION

Face detection is the first step in face recognition system as the output will be used as the input to the face recognition system. One of the methods for face detection is Neural Networks which lies under the category of image based approach. In this paper, we focus on optimizing the RBF Neural Network for face detection. RBFNN is used to distinguish face and non-face. The output of the network can be optimized by setting suitable value of center and spread of the RBF.

II. MOTIVATION TO THE WORK

*A. Radial Basis Function Neural Network*

RBFNN offer several advantages compared to the Multilayer Perceptrons. Two of these advantages are:

- They can be trained using fast 2 stages training algorithm without the need for time consuming non-linear optimization techniques.
- ANN RBF possesses the property of 'best approximation' [8].

RBFNN had been successfully used in face detection such as in [6]. Figure 1 illustrates the architecture of RBFNN used in this work.

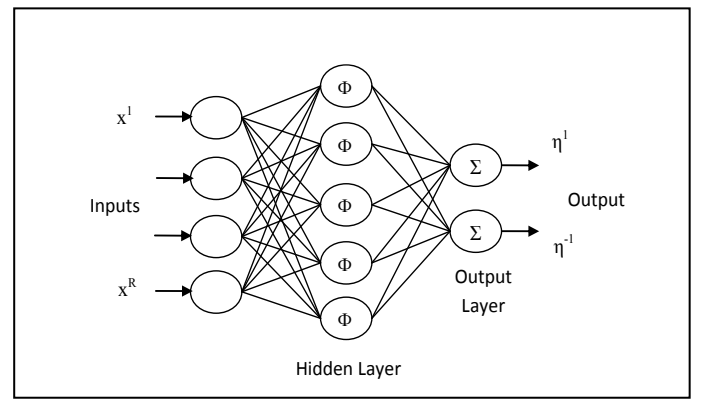

Figure 1: RBF Neural Network

The network consists of three layers: an input layer, a hidden layer and an output layer. Here, *R* denotes the number of inputs while *Q* the number of outputs. For *Q = 1*, the output of the RBF NN in Figure 1 is calculated according to

$$\eta(x,w) = \sum_{k=1}^{S1} w_{1k}\phi(\| x - c_k \|)_2 \qquad (1)$$

where $x \in \Re^{Rx1}$ is an input vector, $\phi(.)$ is a basis function, $\|\cdot\|$ denotes the Euclidean norm, $w_{1k}$ are the weights in the output layer, *S1* is the number of neurons ( and centers) in the hidden layer and $c_k \in \Re^{Rx1}$ are the RBF centers in the input vector space. Equation (1) can also be written as

$$\eta(x,w) = \phi^T(x)w \qquad (2)$$

where

$$\phi^T(x) = [\phi(\| x - c_1 \|) \ldots \ldots \phi_{S1}(\| x - c_{S1} \|)] \qquad (3)$$

and

$$w^T = [w_{11} w_{12} \ldots w_{1S1}] \qquad (4)$$

The output of the neuron in a hidden layer is a nonlinear function of the distance given by:

$$\phi(x) = e^{\frac{-x^2}{\beta^2}} \qquad (5)$$

where β is the spread parameter of the RBF. For training, the least squares formula was used to find the second layer weights while the centers are set using the available data samples.

## III. METHODOLOGY

### A. Network Training

The network is trained using 2429 face data and 4548 non-face data from the CBCL (Center For Biological and Computation Learning) train datasets [5].

K-means clustering is one of the techniques that used to find a set of centers where the technique is more accurately reflects the distribution of the data points. [4] It is used in research such as [6] and [10]. In k-means clustering, the number of desired centers, K, must be decided in advance.

The simplest procedure for selecting the basis function centers $c_k$ is to set the center equal to the input vectors or a random subset of the input vectors from the training set but this is not an optimal procedure since it leads to the use of unnecessarily large number of basis function. [4] Broomhead *et al.* [11] suggested strategies for selecting RBF centers randomly from the training data. The centers of RBF can either be distributed uniformly within the region of input space for which there is data. As for the spread value, we are using the same value of spread for all centers.

For the training, supervised learning is used where training patterns are provided to the RBFNN together with a teaching signal or target. As for the input of face will be given the value of 1 while the input of non-face will be given the value of -1.

### B. Testing

In this experiments, 999 face data and 899 non-face data taken from the CBCL train datasets used as the input. Different centers are chosen ranging from 2 to 200 with the spread value from 1 to 40. Apart from the previous testing, the system will also detect many faces in large image. Image for the testing purpose is taken from [9].

As for the result, the ultimate measure of utility of a biometric system for a particular application is recognition or detection rate [12]. This can be described by two values that are False Acceptance Rate (FAR) and False Rejection Rate (FRR). FAR classifies face data as non-face data while FRR classifies non-face data as face data.

Detection rate that used in this project,

$$\text{Detection Rate} = \frac{\text{number of correct detection}}{\text{total number of input}} \qquad (6)$$

Apart from the previous testing, the system will also detect many faces in a large image. The image for testing is taken from [9]. Sliding window will run inside the image and identified whether there is a face inside the current window. In [9] they are using new reduced set method for Support Vector Machines (SVMs).

## IV. RESULTS

Figure 2, 3 and 4 shows the result for using center 2. In Figure 2, the face detection rate is more than 44% while for non-face is more than 82%. The detection rate will maintain at 64% and 82% for face and non-face as the spread value increases. In Figure 3 we can see that the RBF cannot discriminate face and non-face smoothly. As in Figure 4, the FAR had the lowest value =0.4 while FRR gives the lowest value =0.14. We can see that for the center value 2, the detection rate is very poor.

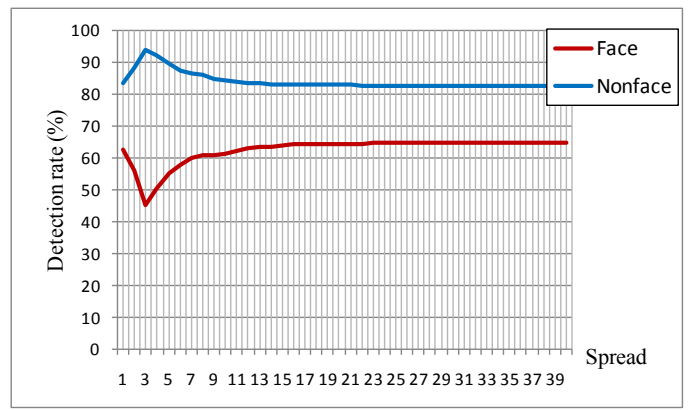

Figure 2: Spread effect on Face & Non-face detection performance – Centre 2

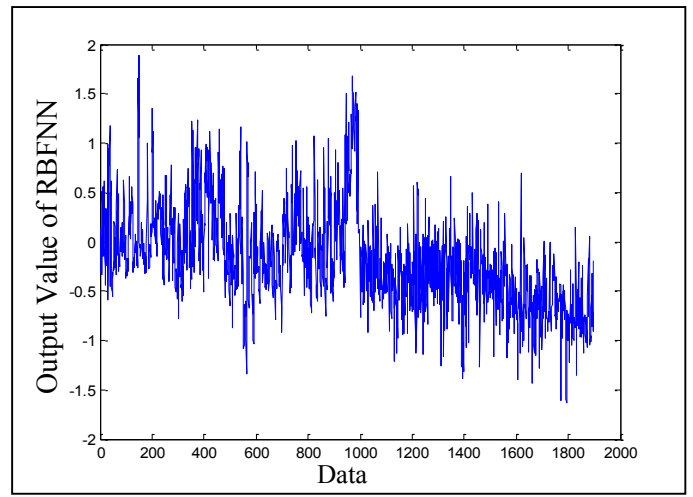

Figure 3: Output response of the trained RBFNN for CBCL training datasets as input for Center- 2/Spread -4

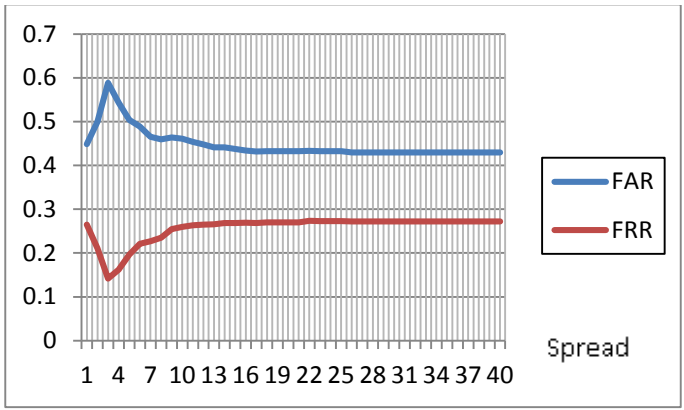

Figure 4: FAR and FRR of the system using center 2

Figure 5, 6 and 7 shows the result of for using center 25. In Figure 5, the detection rate for face and non-face are more than 91%. The detection rate will maintain at 93% for both face and non-face as the spread value increases. Figure 6 shows that the RBF discriminate faces and non-faces data. In Figure 7, the FAR had the lowest value at spread =1 while FRR still gives the lowest value at spread = 4. We can see that for the spread value of 3, the best average detection rate is achieved. The result for center 25 and spread 3 are face detection rate = 93%, non-face detection rate = 95%, FAR = 0.062 and FRR = 0.042.

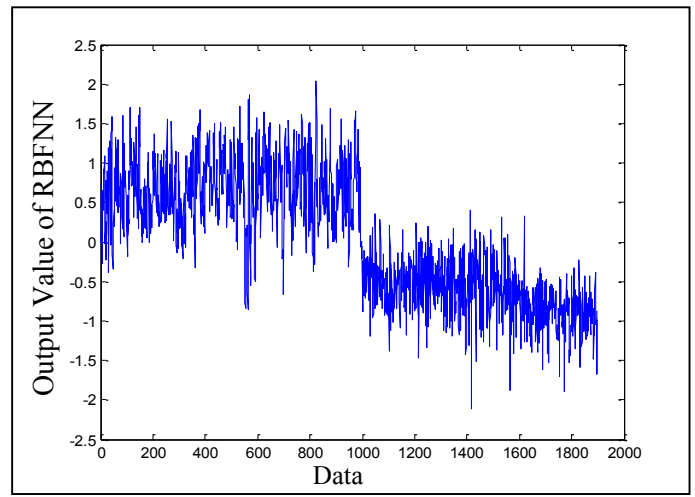

Figure 6: Output response of the trained RBFNN for CBCL training datasets as input for Center- 25/Spread -4

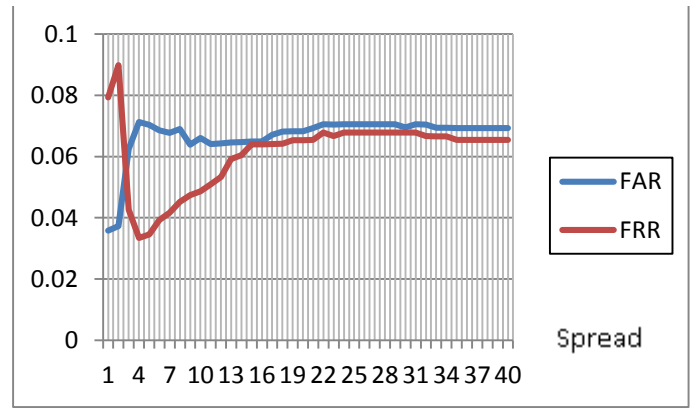

Figure 7: FAR and FRR of the system using center 25

Figure 8, 9 and 10 shows the result of for center 120. In Figure 8, the detection rate for face and non-face are more than 91%. The detection rate will maintain at 95% for both face and non-face as the spread value increases. Figure 9 shows that the RBF response to face and non-face data less discriminative results compare to RBF with center 200. In Figure 10, the FAR and FRR gives the lowest value at spread = 4 before the values increases. We can see that for the spread value of 4, the best detection rate is achieved. The result for center 120 and spread 4 are face detection rate = 95%, non-face detection rate = 98%, FAR = 0.047 and FRR = 0.010.

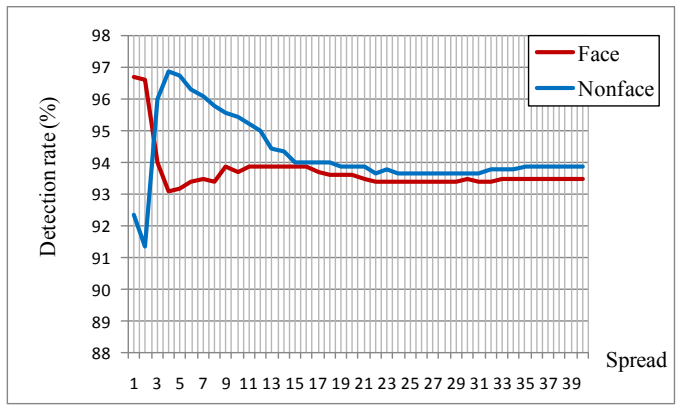

Figure 5: Spread effect on Face & Non-face detection performance – Centre 25

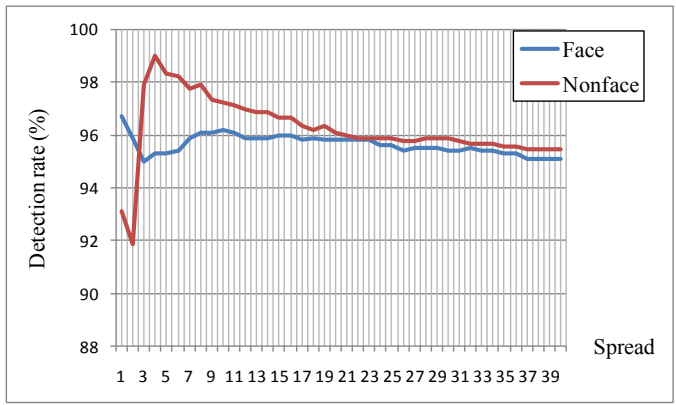

Figure 8: Spread effect on Face & Non-face detection performance – Centre 120

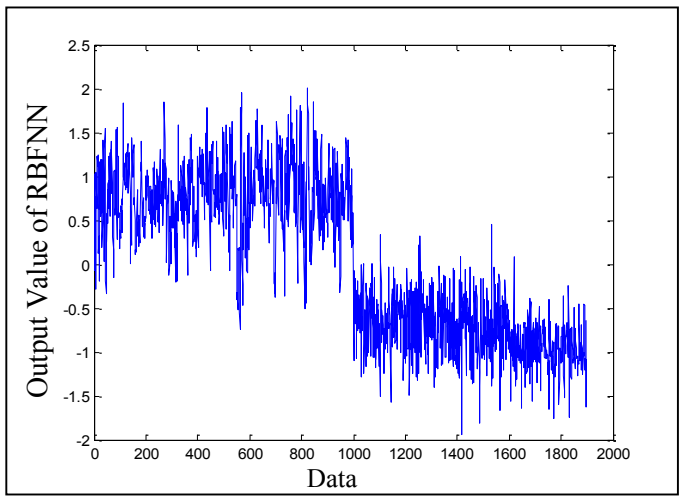

Figure 9: Output response of the trained RBFNN for CBCL training datasets as input for Center- 120/Spread -4

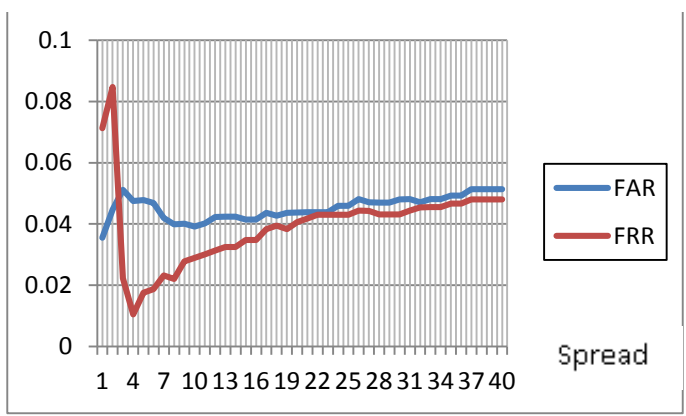

Figure 10: FAR and FRR of the system using center 120

Figure 11, 12 and 13 shows the result of for center 200. In Figure 11, the detection rate for face and non-face are more than 96%. The detection rate will maintain at 95% and 96% for face and non-face as the spread value increases. Figure 12 shows that the RBF response to face and non-face data nicely where it can discriminate faces and non-faces even though not completely. In Figure 13, the FAR and FRR gives the lowest value at spread = 4. We can see that for the spread value of 4, the best detection rate is achieved. The result for center 200 and spread 4 are face detection rate = 97%, non-face detection rate = 99%, FAR = 0.029 and FRR = 0.004.

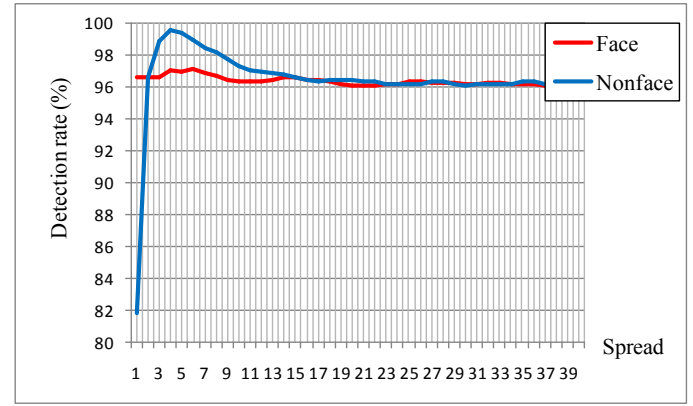

Figure 11: Spread effect on Face & Non-face detection performance – Centre 200

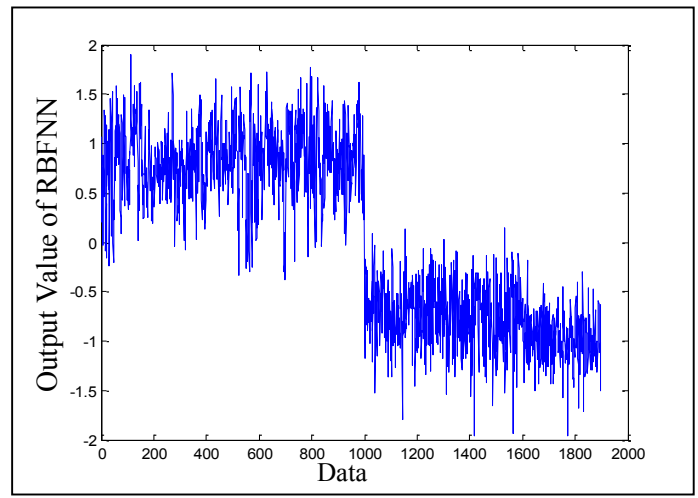

Figure 12: Output response of the trained RBFNN for CBCL training datasets as input for Center- 200/Spread -4

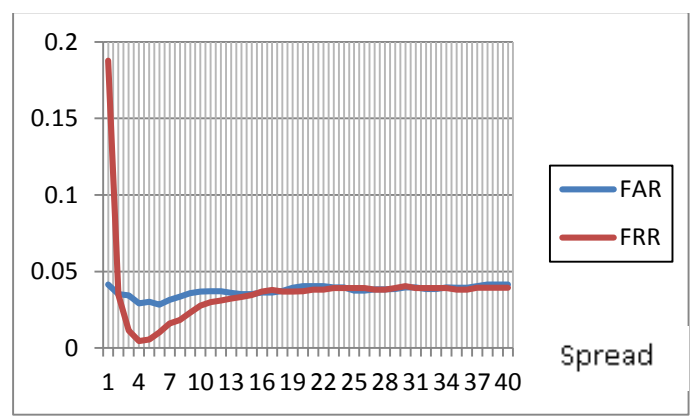

Figure 13: FAR and FRR of the system using center 200

For detecting many faces in an image, setting the system with center 2 as shown in Figure 14 gives the worst result as there are too many false negative. It is hard to see whether the system detect all faces as there are too many overlapped rectangular. Increasing the value of center improved the result as the false negative becomes less as in Figure 15. For the use of center 5 and spread 10, the system cannot detect two faces and yet have two false negative.

Starting with center 25 and above, the results are much better as the system can detect all faces in the picture without any false negative as we refer to Figure 16. But still the spread value is crucial where we can still get false negative as shown in Figure 18 and Figure 20 for center 25 and 200.

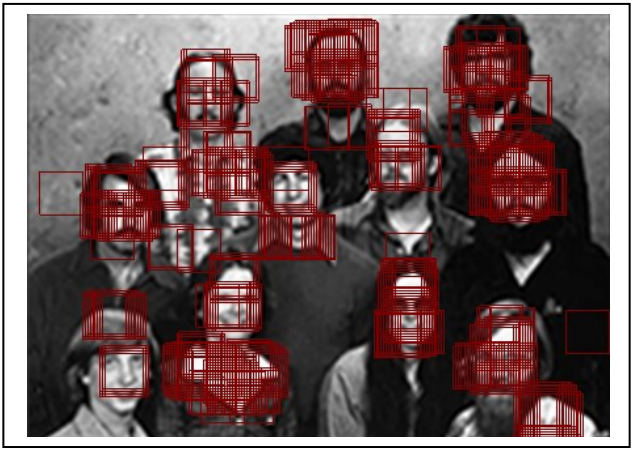

Figure 14: Using RBF with the value of centre 2 and spread 4

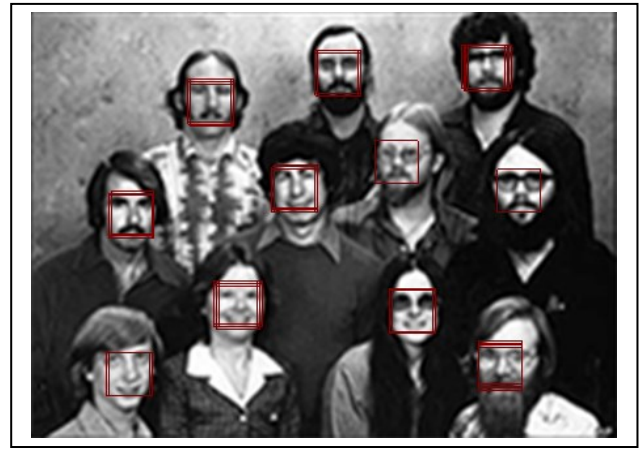

Figure 17: Using RBF with the value of centre 25 and spread 5

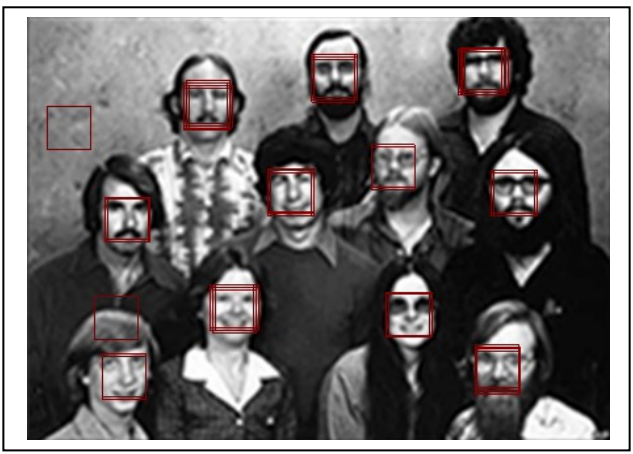

Figure 15: Using RBF with the value of centre 5 and spread 4

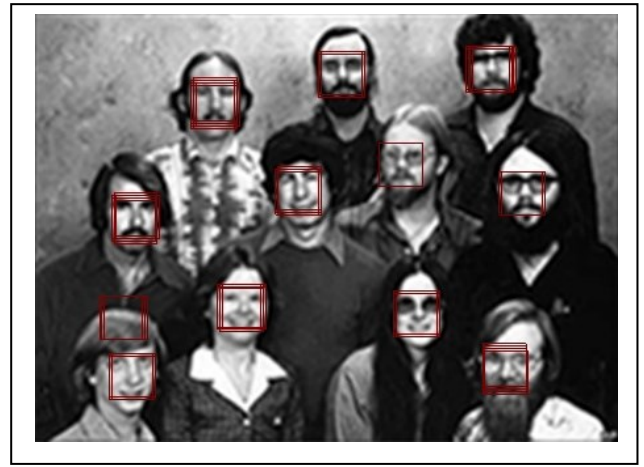

Figure 18: Using RBF with the value of centre 25 and spread 4

For Center 200, spread value of 5 will ensure that the system detect all faces in the image without false negative as shown in Figure 19. As for spread 4, the system can detect all faces but with 2 false negative. If we refer to Table 1, the system should have the best setting at spread 4 as the FRR is low that is 0.004. This maybe because the non-face image that had been classified as face are in the range of 0.004 FRR.

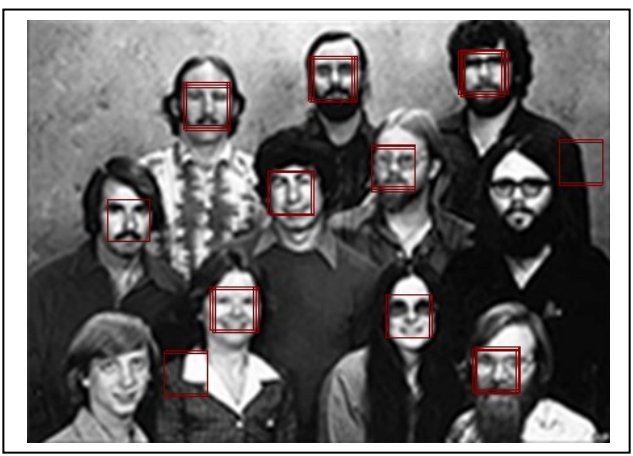

Figure 16: Using RBF with the value of centre 5 and spread 10

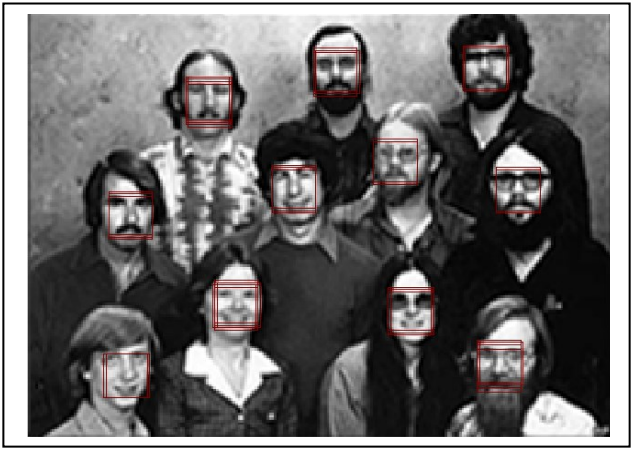

Figure 19: Using RBF with the value of centre 200 and spread 5

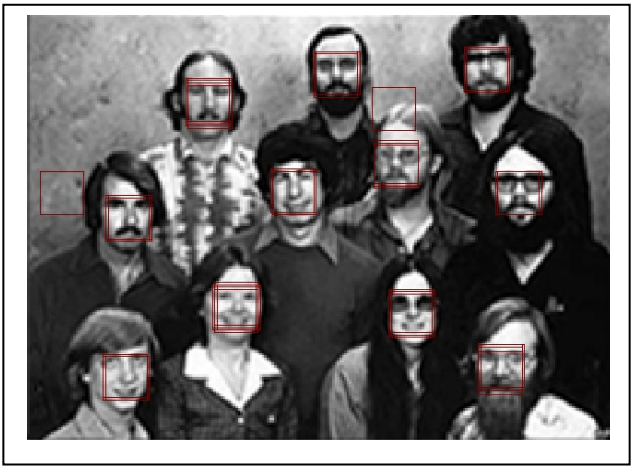

Figure 20: Using RBF with the value of centre 200 and spread 4

Table 1: Summary of the best setting for RBF Face Detection

| Centre: | Spread | Detection Rate for Face | Detection Rate for Nonface | FAR | FRR |
|---|---|---|---|---|---|
| 200 | 4 | 97.0971 | 99.55506 | 0.029159 | 0.004582 |
| 200 | 5 | 96.997 | 99.44383 | 0.030198 | 0.005734 |
| 120 | 4 | 95.2953 | 98.99889 | 0.047523 | 0.010505 |
| 120 | 5 | 95.2953 | 98.33148 | 0.047845 | 0.017509 |
| 100 | 4 | 96.3964 | 98.55395 | 0.036565 | 0.015001 |
| 100 | 5 | 96.1962 | 98.33148 | 0.038683 | 0.017345 |
| 25 | 4 | 93.093 | 96.885 | 0.071291 | 0.033461 |
| 25 | 5 | 93.193 | 96.774 | 0.070339 | 0.034616 |
| 20 | 4 | 95.495 | 93.326 | 0.048272 | 0.069888 |
| 20 | 5 | 94.795 | 94.438 | 0.055116 | 0.058674 |

## V. DISCUSSIONS

For the results with different value of spread applied, the results show that spread value of 4 and 5 gives the best result for any centers chosen. Even though the performance increases and maintain at certain level as spread becomes larger, the setting of spread larger than 40 is impractical. Most of the FAR and FRR also have low value at spread 4 and 5. If we consider the lowest FAR and FRR, center 200 and spread 4 is the best setting. But for detecting faces in the image taken from [9], the result still gives two false negative. In section IV, the setting that can detect all faces in the image is center 200 with spread 5 and center 25 with spread 5. Among these setting, Center 200 and Spread 5 give the best result in term of face detection rate, that is low FAR and FRR as well as can detect all faces in the image taken from [9] without false negative. The response of the network using center 200 also shows that it gives more discriminative results compare to other setting. As the center chosen is larger, the time for training the network will increase. Using this setting that is center 200 and spread 4 instead of using all input data as the center is fair enough. This is because it took about half an hour for training the RBFNN using center 200. Using all the datasets as input as discussed in section III that contains more than six thousand data might take days for the program to settle.